\documentclass[sigconf]{acmart}

\setcitestyle{nosort}

\usepackage{booktabs} 
\usepackage{algpseudocode}
\usepackage{pifont}
\usepackage{pgfplots}
\usetikzlibrary{shapes,arrows,automata}
\usepackage{amsmath,amsthm,amssymb}
\usepackage{tikz}
\usetikzlibrary{arrows,automata}
\usepackage{caption}
\usepackage{subcaption}
\usepackage{makecell}
\usepackage{float}
\usepackage{enumitem}
\usepackage[linesnumbered,ruled]{algorithm2e}
\usepackage{color}
\usepackage{hyperref}
\usepackage{cleveref}
\usepackage{soul}

\setcopyright{rightsretained}
\newcommand{\ie}{\textit{i.e., }}
\newcommand{\eg}{\textit{e.g., }}

\acmDOI{10.1145/nnnnnnn.nnnnnnn}

\acmISBN{978-x-xxxx-xxxx-x/YY/MM}

\acmConference[GECCO '19]{the Genetic and Evolutionary Computation Conference 2019}{July 13--17, 2019}{Prague, Czech Republic}
\acmYear{2019}
\copyrightyear{2019}

\acmPrice{15.00}

\begin{document}
\title[Preference-based Multiobj. VMP: A Ceteris Paribus Approach]{Preference-based Multiobjective Virtual Machine Placement: A Ceteris Paribus Approach}

\author{Abdulaziz Alashaikh}
\orcid{0000-0002-1525-4500}
\affiliation{%
  \institution{}
  \department{Umm Al-Qura University}
  \city{Makkah, Saudi Arabia} 
}
\email{azizoozi@gmail.com}

\author{Eisa Alanazi }
\affiliation{%
  \institution{Dept. of Computer Science}
  \department{Umm Al-Qura University}
  \city{Makkah, Saudi Arabia} 
}
\email{eaanazi@uqu.edu.sa}

\begin{abstract}
This work adopts the notion of Ceteris Paribus (CP) as an interpretation of the Decision Maker (DM) preferences and incorporates it in a constrained multiobjective problem known as virtual machine placement (VMP). VMP is an essential multiobjective problem in the design and operation of cloud data centers concerned about placing each virtual machine to a physical machine (a server) in the data center. We analyze the effectiveness of CP interpretation on VMP problems and propose an NSGA-II variant with which preferred solutions are returned at almost no extra time cost. 
\end{abstract}

%
%




\maketitle

\section{Introduction}

Recent years have shown an increased interest in preference-based evolutionary algorithms for multiobjective optimization. However, investigating the Decision Maker (DM) interpretation of preferences has received little attention. Such interpretation is crucial to return solutions that are plausible to the DM. To this end, the 
\emph{Ceteris Paribus} (CP) interpretation of preferences is believed to be naturally exercised by many DMs \cite{hansson1996ceteris}. In multiattribute domains, it is usually meant to be when all other attributes are fixed. Under the CP semantics, a DM stating a preference for $\alpha$ over $\beta$ means for any two solutions $s_1$ and $s_2$, the solution with $X=\alpha$ is preferred to another with $X=\beta$ \emph{only} when $s_1$ and $s_2$ share same values for other variables (assuming $\alpha$ and $\beta$ are two possible values of the variable $X$). In this work, we investigate the applicability of CP preferences via a concrete constrained Evolutionary Multiobjective Optimization (EMO) problem known as Virtual Machine Placement (VMP) where the CP interpretation is applied to handle the DM preferences on the decision space. 
VMP has been tackled in the literature via different EMO approaches \cite{Attaoui2018,Gao2013r,Xu2010r}. However, to the best of our knowledge, the notion of ceteris paribus preferences has not been investigated yet. Our experiments suggest that our approach correctly finds Pareto solutions that reflect DM's preferences. 

\section{Ceteris Paribus  Preferences}
\label{sec:CPsem}
Let $V=\{V_1,V_2,\dots,V_n\}$ be a set of decision variables where every variable $V_i\in V$ has a set of possible values (\ie its domain) $dom(V_i)$. A preference is simply an irreflexive, transitive relation $\succ$. We use $\succ_i$ to refer to the preference of variable $V_i\in V$ defined over $dom(V_i)$ and $\alpha\succ_i \beta$ to mean $\alpha$ is preferred over $\beta$ where $\alpha,\beta\in dom(V_i)$. A solution $s$ is a mapping for every variable $V_i\in V$ to a value from $dom(V_i)$, and the set of all solutions is denoted by $\mathcal{O}$. We use $s[Y]$ to denote the projection of the assignment $s\in \mathcal{O}$ to $Y\subset V$.

\begin{figure*}[t]
\centering
\begin{minipage}[]{0.35\textwidth}
\includegraphics[width=.9\textwidth]{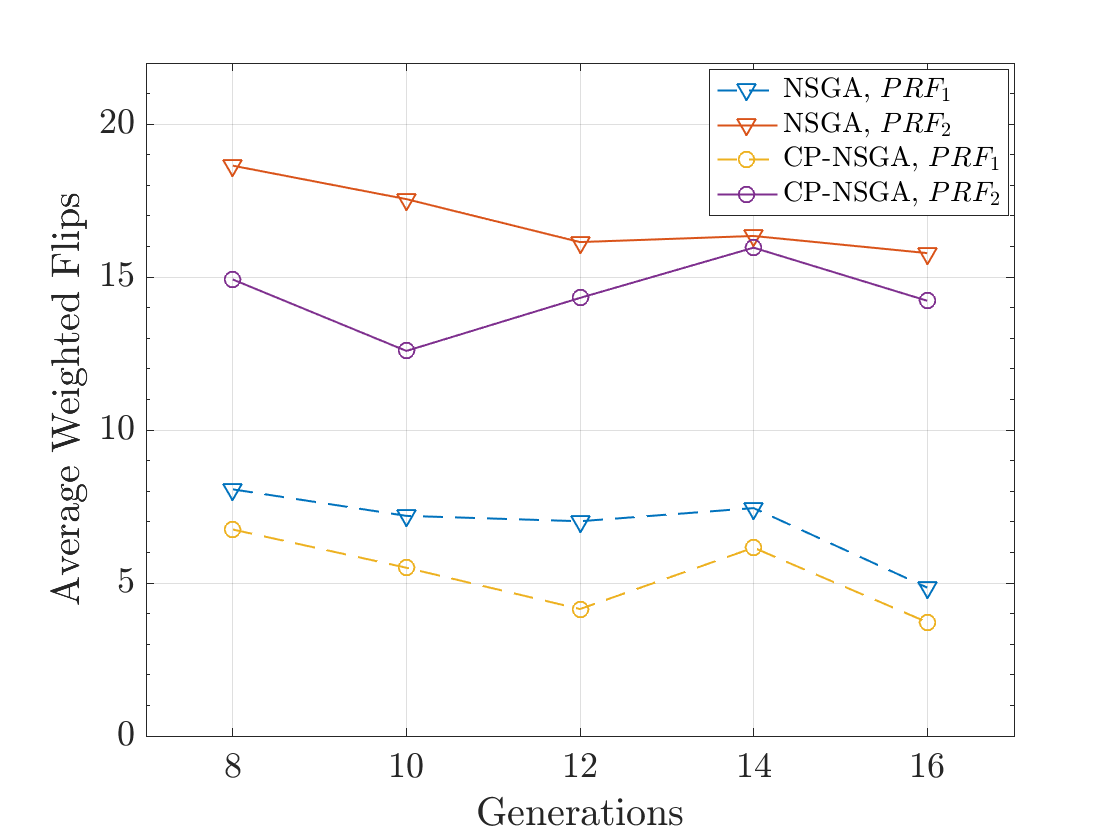}
\caption{Weighted flips comparison.}
\label{fig:nsga_cpnsga}
\end{minipage}\hfill
  \centering
  \begin{minipage}[]{0.65\textwidth}
  \begin{subfigure}[]{.5\textwidth}
    \centering\includegraphics[width=.9\textwidth]{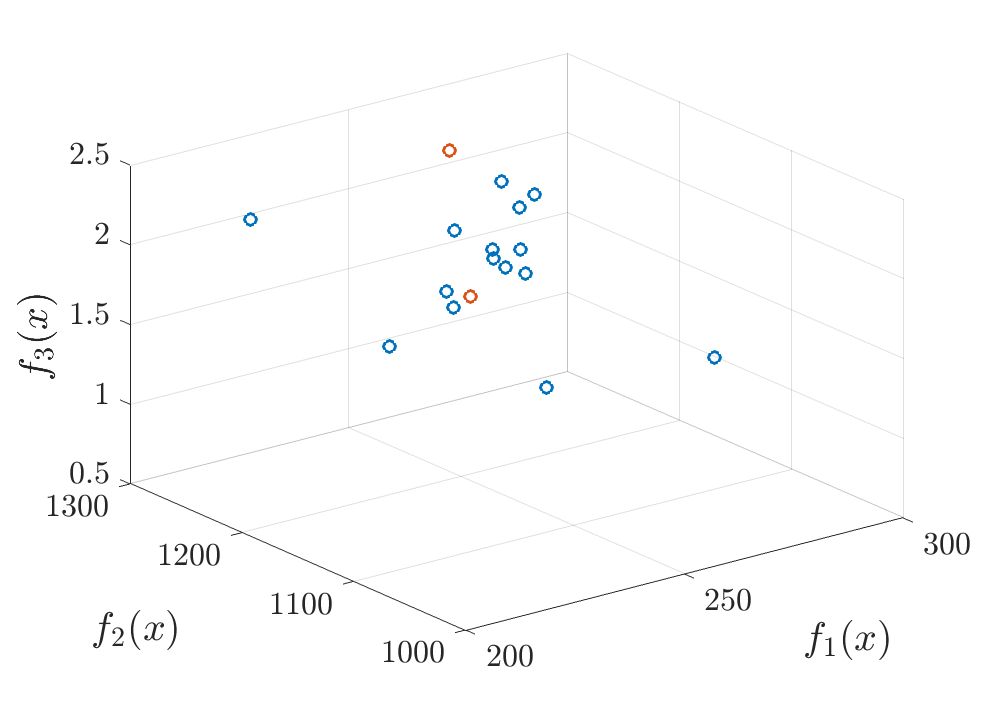}
    \caption{8 generations}\label{fig:6_18_prf1_ev8}
    \end{subfigure}\hfill%
    \begin{subfigure}[]{.5\textwidth}
    \centering\includegraphics[width=.9\textwidth]{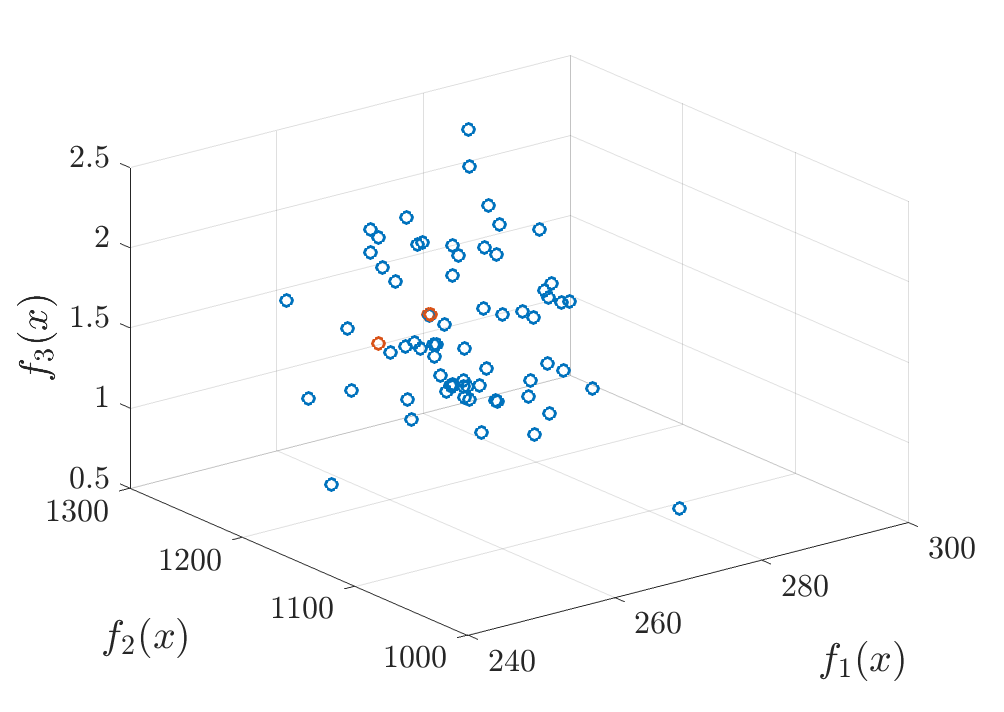}
    \caption{16 generations}\label{fig:6_18_prf1_ev16}
    \end{subfigure}%
    \caption{PF of different generations for $PRF_1$.}\label{fig:6_18_prf_comp}
\end{minipage}
\end{figure*} 

An attractive structure in this regard is the separable Ceteris Paribus (SCP) structure \cite{lang2009complexity}. SCP is defined over $V$ by supplying a preference statement $\succ_i$ for every $v_i\in V$ where the dominance relation in SCPs is defined as follows: For any two solutions $s,s'\in \mathcal{O}$, let  $Diff(s,s')=\{Y\in V~|~s[Y]\neq s'[Y]\}$ be the set of variables with different values in $s$ and $s'$. We say $s$ dominates $s'$ (i.e., $s\succ s'$) if and only if $s[V_i]\succ_i s'[V_i]$ for every $V_i\in Diff(s,s')$\cite{lang2009complexity}. SCPs are computationally attractive in a sense that the dominance relation between any two solutions can be determined in linear time. 
 Notice that if $s$ does not dominate $s'$, it is not necessarily that it is the case where $s'\succ s$ as both can be incomparable. We can then define the notion of Pareto optimality for a solution as follows: a solution $s\in \mathcal{O}$ is said to be CPR-Pareto (from Ceteris Paribus Rules) if and only if there exists no solution dominates it. In constrained EMO problems, the feasibility  as well as being Pareto in the objective space  are required as well.

\section{Virtual Machine Placement with Ceteris Paribus}
 \label{sec:VMP}
The virtual machine placement (VMP) problem is an essential problem in the design and operation of cloud data centers.  
The VMP problem is defined as follows: Given a set of physical machines and  a set of virtual machines, how to place each virtual machine (VM) into a physical machine (PM) while optimizing a set of objectives and satisfying a set of constraints. 
Each VM requires some resources described in terms of CPU cycles and RAM space. 
While a server $P_k$ can host multiple VMs, the resource utilization (\ie load) should not exceed its capacity, which is the CPU and memory capacities. 
The objective is to minimize 
the total communication cost ($C_{V_i,P_k}^{ V_j, P_l}$) \cite{Shrivastava2011r},  
the total power ${E_k}$ \cite{Xu2010r}, 
and the resource wastage index \cite{Gao2013r}. 

We propose a variant of the NSGA-II  (CP-NSGA) that takes into consideration the CP information issued by DM during the search by introducing a new CPR operator $\triangleleft$ that favors solutions that are close to the DM preferences. We keep the non-dominated sorting over the objective space intact to some extent while incorporating those solutions that are promising in terms of meeting the DM preferences. 
Initially, we assume an SCP structure $\mathcal{N}$ is provided along with the multiobjective problem. Then, after sorting the nondominated solutions in NSGA-II, $\triangleleft$ takes the elements of the last rank and fed them to dominance relation to identify who is CPR-Pareto w.r.t~$\mathcal{N}$. 
Therefore, our algorithm resembles the NSGA-II algorithm except that we embed the CPR operator in the next generation selection phase in front of the crowding distance operator.


\section{Experiments}
\label{sec:experiments}
%
%
A preference for the $i$th virtual machine (denoted as $PRF(i)$) is simply a random permutation of $\{1,...,m\}$. That is, if $PRF(i)$=\{2,3,1\} then the most preferred placement of $V_i$ is in $P_3$, $P_1$, and then in $P_2$. 
If a problem has $k$ VM with preferences, then the score of a solution will be a vector of $k$ scores. Those scores form the solutions of the $k$ decision variables that have preferences. Then, the scores of multiple solutions will be the input for CPR operator. 

 We consider two scenarios $PRF_1$ and $PRF_2$ with 6 PMs and 8 VMs and number of preferences equal to three and six respectively.  Both scenarios were solved using different settings for number of generations and the population size was set to 100. 

\Cref{fig:nsga_cpnsga} compares the weighted flips of NSGA-II and CP-NSGA Pareto of the two scenarios  averaged over 5 repetitions. These generations are of independent runs.
The weighted flips is the number of flips (or changes) required for a placement decision to reach the optimal placement. The optimal placement is the one where every VM is placed into its best physical machine. 
For example, If the PRF score for 3 VMs is \{1, 3, 5\}, it means that the first VM was placed in its most preferred PM, whereas second and third VMs were placed at the third and fifth preferred PM. The weighted flips is (0+3+5=8). 
We can see that CP-NSGA score are lower than NSGA. 
%

%
\Cref{fig:6_18_prf_comp} shows a sample result for the evolution of $PRF_1$  solutions where the Pareto Front solutions are blue and CPR-Pareto solutions are red. We notice that as the PF evolves, the CPR ranking evolves. 
While the CP-NSGA algorithm persistently tries to keep its solutions, the CPR ranking changes with new generations as it would have a chance to measure the preference of 
a previously higher ranked Parent evaluated at a lower rank in the subsequent iterations, or measure the preference of a newly generated offspring. 
%
As the number of generations increases, the possibility of finding better CPR-Pareto increases, \eg the case in \Cref{fig:6_18_prf1_ev16} reached the optimal preferences for the 3 VMs. 
Also, we find the algorithm can successfully pass CPR-Pareto solutions to the final population without degrading the quality of the Pareto solutions. 
For example, in scenario-1 \{34\%, 39\%, 31\%, 28\%, 21\%\} of the last ranked individuals were chosen by the CPR, 
compared to \{94\%, 86\%, 78\%, 87\%, 90\} for scenario-2 with larger $\#PRF$.
%
%
Lastly, the run time for these instances on CP-NSGA has almost negligible increase in comparison with NSGA. 
%
%

\section{Concluding Remarks}
\label{sec:conclusions}
We have proposed a variant of NSGA-II with which we can promote solutions that are promising in terms of CP to the next generations. This is done by incorporating CPR operator into the selection phase of NSGA-II. In essence, our algorithm tries to pass as many CPR-Pareto as possible to the final generation. Our preliminary results show that our variant still respects diversity to some extent while outputting preferred solutions at almost no extra time cost.

\begin{acks}
This work is supported by the Postdoctoral Initiative at the R\&D Office, Ministry of Education, Saudi Arabia.
\end{acks}

\bibliographystyle{ACM-Reference-Format}

\bibliography{ref.bib} 

\end{document}